# System Dynamics Modelling of the Processes Involving the Maintenance of the Naive T Cell Repertoire

Grazziela P. Figueredo
IMA Research Group School
of Computer Science
University of Nottingham
Nottingham, NG8 1BB
UK
gzf@cs.nott.ac.uk

Uwe Aickelin
IMA Research Group School
of Computer Science
University of Nottingham
Nottingham, NG8 1BB
UK
uxa@cs.nott.ac.uk

Amanda Whitbrook
IMA Research Group School
of Computer Science
University of Nottingham
Nottingham, NG8 1BB
UK
amw@cs.nott.ac.uk

## Abstract

The study of immune system aging, i.e. immunosenescence, is a relatively new research topic. It deals with understanding the processes of immuno-degradation that indicate signs of functionality loss possibly leading to death. Even though it is not possible to prevent immunosenescence, there is great benefit in comprehending its causes, which may help to reverse some of the damage done and thus improve life expectancy. One of the main factors influencing the process of immunosenescence is the number and phenotypical variety of naive T cells in an individual. This work presents a review of immunosenescence, proposes system dynamics modelling of the processes involving the maintenance of the naive T cell repertoire and presents some preliminary results.

## 1 Introduction

The study of aging in human beings is a relatively new research topic and deals with understanding the processes of tissue degeneration so that they can be stopped or slowed down. Some stages of degradation indicate signs of functionality loss that precede the end of life. Even though it is not possible to prevent aging, there is great benefit in understanding its causes, which may help to reverse some of the damage done and thus improve life expectancy.

This work is concerned with the aging of the immune system (IS), known as immunosenescence. The IS acts to fight and prevent many kinds of diseases in an individual throughout life. However, with age there is a decay of the IS performance resulting in degenerative diseases, deregulated and ineffective immune responses. This can end up in complete collapse of the defence mechanisms resulting in death. Various theories have been proposed to explain this phenomenon, including the levels of antigenic stress, oxidation, lack of cellular resources and DNA damage. In order to investigate how the IS ages and fails it is necessary to understand the processes by which the instabilities take place, develop, propagate and turn out to be destructive.

By acquiring such knowledge it would be possible to create predictive models that can estimate the immune fitness of a patient at a certain age given their vaccination history, exposure to diseases and hemograms. Moreover, the knowledge could be applied to any other aging process, since there are many scientific, social, security, engineering, economical and computationally important problems demanding designs with more predictable degenerative associated properties.

A degenerative system is one that, through a series of sequential events, devolves in time until it reaches a point where its functionality is compromised. Trying to predict when a certain system will no longer work properly to satisfy its original purpose has great benefit because the system can be replaced, upgraded or redesigned before this point is reached. This has potential implications for the safety, security and welfare of resources such as water, air transport, energy, product quality, computer networks and control.

There are many factors that can cause a system to deteriorate. From the hardware interactions perspective aging refers to progressive performance degradation or a sudden hang or crash of a system due to exhaustion of operating system resources and accumulation of problems. Another cause is the negligence of a system's owners to modify it to meet changing needs and new demands in the market, such as new sources of information, new processes and technology upgrades. Other age-associated problems can occur as the result of changes that are made without proper testing, introducing new failures. Sometimes systems manufacturers find it difficult to keep up with the market and to compete with new products. This work presents a review of the immunosenescence phenomenon and includes a description of the various theories that could be used for modelling and developing algorithms for solving degeneration problems. Also, a proposal for a first simulation model and some preliminary results are presented.

## 2 Immunosenescence

According to Bulati et al. [2], aging is a complex process that negatively impacts the development of the immune system and its ability to function. Progressive changes in the innate and adaptive immune systems have a major impact on the capacity of an individual to produce effective immune responses. These changes that characterize the aging of the immune system are called immunosenescence. The decrease of immunocompetence in the elderly can be envisaged as the result of the continuous challenge of the unavoidable exposure to a variety of potential antigens, i.e. viruses, bacteria, food and self-antigens [5]. Antigens are the cause of a persistent life-long antigenic stress, responsible for the filling of the immunological space by accumulation of effector T cells and immunological memory [5]. With age, there is also a significant reduction of naive T cells caused by the involution of the thymus. This situation eventually leaves the body more susceptible to infectious and non-infections diseases [8]. In addition, there is evidence that clonotypical immunity deteriorates, while ancestral innate or natural immunity is conserved or even up-regulated [5, 10].

According to Franceschi [5], some factors that characterize immunosenescence are the accumulation of memory T cells, the decrease and exhaustion of naive T cells and a marked reduction of the T cell repertoire. Bulati [2], on the other hand, believes both innate and adaptive immunity are usually involved in the pathogenesis of chronic age-related diseases like arthritis, atherosclerosis, osteoporosis, diabetes, etc. However, innate immunity appears to be the prevalent mechanism driving tissue damages associated with different age-related diseases [2]. Thus, aging is accompanied by an age-dependent up-regulation of the inflammatory response, due to the chronic antigenic stress that impinges throughout life upon innate immunity, and has potential implications for the onset of inflammatory diseases. Bulati points out some important factors related to aging:

- Reactivity of dendritic cells to self antigens can be characteristic of aging. Furthermore, this over-reactivity induces T lymphocyte proliferation with subsequent higher risk of autoimmune diseases.

- Hyper activated T cells are possibly involved in bone loss associated with vascular disease in aged mice.

- There is a decrease in vaccine responsiveness.

De Martinis [6] and Franceschi [5] state that the most important characteristics of immunosenescence are the accumulation of memory and effector T cells, reduction of naive T cells, shrinkage of the T cell repertoire and a filling of immunological space. They point out that:

- The filling of immunological space with memory and effector cells is a consequence of exposure to a variety of antigens over time.

- Clonal expansion of peripheral T cells carrying receptors for single epitopes of both herpes virus Cytomegalovirus (CMV) and Epstein-Barr virus (EBV) are common in the elderly and are associated with the loss of early memory cells, an increase of T cytotoxic cells and a gradual filling of immunological space.

- With the decline of immune function there is an increase in autoantibody frequency. An important result of this may be the loss of ability to distinguish between self and nonself molecules.

- The lifelong respiratory burst, i.e. reactive oxygen species (ROS) causes damage to important cellular components (lipidic membranes, enzymatic and structural proteins and nucleic acids) during aging. ROS damage is counteracted by several genetically controlled enzymatic and non-enzymatic antioxidant defense systems. All these protective mechanisms tend to become less effective with age.

- An elderly IS becomes more predisposed to chronic inflammatory reactions and less able to respond to acute and massive challenges by new antigens. Inflamm-aging, the peculiar chronic inflammatory status which characterizes aging, is under genetic control and is detrimental to longevity. It leads to long term tissue damage and is related to an increased mortality risk.

- The chronic overexposure to stressors determines a highly pathogenic sustained activation of the stress-response system leading to a progressively reduced capacity to recover from stress-induced modifications.

## 3 Evolutionary-based Immunosenescence

De Martinis [6] and Franceschi [5] state that human immunosenescence can be envisaged as a situation in which the most evolutionary recent, sophisticated defense mechanisms deteriorate with age, while the most evolutionary old and gross mechanisms are preserved or are negligly affected and, in some cases, up-regulated. Therefore, the aging of the immune system is not a random process without rules or directions, but rather is subject to evolutionary constraints. The antagonistic pleiotropy theory of aging [4] states that natural selection has favoured genes conferring short-term benefits to the organism at the cost of deterioration in later life. Evolutionary ideas lead to the belief that the IS has been selected to serve individuals only until reproduction ceases. This means that the optimum functioning of the IS is only guaranteed for the number of years in which an individual is capable of reproduction [4]. After that, all the biochemical processes proceed freely without any past selective pressure to improve the life of that individual. The trend of thymic ontogenesis and involution in early age supports this hypothesis. Humans had a life expectancy of between 30 and 50 years two centuries ago. Nowadays the IS must serve individuals living to between 80 and 120 years, which is much longer than predicted by evolutionary forces. Therefore, old people have to cope with a lifelong antigenic burden encompassing several decades of evolutionary unpredicted exposure. This chronic antigenic stress and subsequent inflammatory burden have a major impact on survival and frailty.

## 4 Candidates for Immunosenescence Models

The four most influential theories from the above are selected as possible candidates for building computationally predictive systems. These theories are immunological space filling with memory cells, lack of naive T cells, the innate system up-regulation, and accumulation of T-regulatory cells.

### 4.1 Space Filling

The immune system deteriorates with age by losing functionality and immunocompetent cells. Moreover, it becomes limited in its use of resources; there is a finite number of T cells in operation at any time and to work properly, the IS needs a reserve of naive T cells for new intrusions, and memory cells for previously encountered antigens. With age, the repertoire of naive T cells shrinks proportionately to faced threats, while memory increases [5, 6, 8]. Late in life the T cell population becomes less diverse and some antigen-specific types of T cell clones can grow to a great percentage of the total T cell population, which takes up the space needed for other T cells, resulting in a less diverse and ineffective IS. At some point there are not enough naive T cells left to mount any sort of effective defense and the total repertoire of T lymphocytes is filled with memory cells.

### 4.2 Lack of Naive T Cells

Before 20 years of age the set of naive T cells is sustained primarily from thymic output [8]. However, in middle age there is a change in the source of naive T cells; as the thymus involutes, there is a considerable shrinkage in its T cell output, which means that new T cells are mostly produced by peripheral expansion. There is also a belief that some memory cells have their phenotype reverted back to naive cells [8]. These two new methods of naive T cell repertoire maintenance are not effective [8] as they do not produce new phenotypic changes on lymphocytes. Rather, evidence shows that they keep filling the naive T cell space with copies of existing cells. Therefore, the age-related loss of clones of some antigen-specific T cells could be irreversible, because there are no more naive T cells to maintain these clones. These age-related phenomena lead to a decay of performance in fighting aggressors.

### 4.3 Innate Up-regulation

With age there is a decay in adequate functioning of the main phagocytes, i.e. macrophages, neutrophils [3] and dendritic cells [1]. As a consequence, deregulated immune and inflammatory responses occur in old people. The investigation into the cellular and molecular mechanism underlying these disorders has provided compelling evidence that up-regulation plays a critical role in the age-associated problems of the immune and inflammatory responses [10]. Thus, innate immunity and a high capacity for mounting a strong inflammatory response, useful at a younger age can become detrimental later in life. Inflamm-aging can thus be considered the main phenomenon responsible for major age-related diseases and the evolutionary price to pay for an immune system that is fully capable of defending against infectious diseases earlier in life.

### 4.4 Accumulation of $T_{reg}$ Cells

The individual's ability to mount an effective immune response can be limited by regulatory elements such as significant changes in the number of T regulatory ($T_{reg}$) cells [9]. $T_{reg}$ cells act to suppress activation of the IS and thereby maintain immune system homeostasis and tolerance. They can accumulate or reduce with age. The accumulation of $T_{reg}$ cells in the old inhibits or prevents some immune responses, i.e. anti-tumoral ones. Also, the reduction of $T_{reg}$ cells might compromise the activation of immune responses in the aged. Therefore, an imbalance in $T_{reg}$ normal functioning can predispose immune dysfunction. This results in a higher risk of immune-mediated diseases, cancer or infections.

### 4.5 Discussion

A summary of the main characteristics of the candidate models described in the previous sections is presented in Table 1. The aim of this work is to develop a model that considers each of these characteristics and also define more simple models in order to investigate individual aspects of the immunosenescence phenomenon. In particular, simulation methods such as system dynamics (SD) and agent based simulation (ABS) would be suitable for modelling the phenomena. An SD approach aims to understand the behavior of complex systems over time. It works with feedback loops and stocks and flows that help describe the systems' nonlinearity. ABS is concerned with modelling agents to observe their behaviors given changes to the environment. The model proposed in this paper is based on reference [8] and involves interactions that influence the naive T cell populations. It is described in the next section.

## 5 Model: Naive T Cell Output

Some markers of thymic contribution in an individual throughout life are defined by levels of T cell receptor excision circle (TREC), which is one episomal circular DNA formed during the coding of the T-cell receptor. TREC percentage decays with shrinkage of thymic output and activation of naive T cells. The model proposed here is based on the data and equations obtained in [8], which is concerned with establishing an understanding of naive T cell repertoire dynamics. The model's objective is to determine the likely contribution of each of the naive T cell's sustaining sources, by comparing estimates of TREC (see Figure 1). The dynamics of naive proliferation, TREC and reversal of memory to naive T cells are modelled mathematically. An SD model of the mathematical model described in [8] is proposed in order to reproduce and further investigate the processes involving the maintenance of naive T cell populations. The variables involved are the thymic output rate, the number of active cells, the number of memory cells turning into naive cells and the number of naive T cells obtained by naive peripheral proliferation. The number and phenotypical variety of naive T cells in an individual is one of the main factors influencing the process of immunosenescence. The simulation aims to investigate the interactions between each variable and their relevance to the maintenance of the naive T cell repertoire.

| Theories | | | | |
|---|---|---|---|---|
| Characteristics | 4.1 | 4.2 | 4.3 | 4.4 |
| *Shrinkage of naive cells* | × | × | | |
| *Diversity decrease* | × | × | | |
| *Few clones taking space* | × | × | | |
| *Excessive memory cells* | × | | | |
| *Loss of clones* | | × | | |
| *Inflammation* | | | × | × |
| *Excessive T suppression* | | | × | × |
| *Degeneration* | × | × | × | × |
| *Auto-immunity* | × | × | × | × |
| *Less vaccine response* | × | × | | × |

Table 1: Main characteristics of the candidate models.

### 5.1 The Mathematical Model

The mathematical model proposed in [8] is described by the differential equations (1) to (3) below, where $N$ is the total number of naive cells of direct thymic origin, $N_p$ is the number of naive cells that have undergone proliferation, $A$ is the number of activated cells, $M$ is the number of memory cells and $t$ is the time (in years). The first differential equation is:

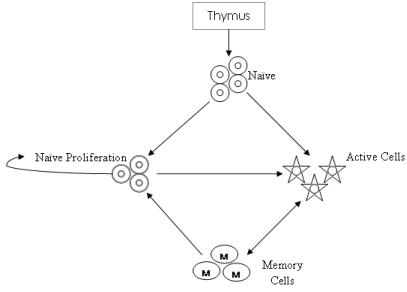

Figure 1: Dynamics of naive T cells.

$$\frac{dN}{dt} = s_0 e^{-\lambda_t t} s(N_p) - [\lambda_n + \mu_n g(N_p)]N, \quad (1)$$

where $s_0$ is the thymic output (equal to 1.65); $\lambda_t$ is the thymic decay rate (equal to $\frac{log(2)}{15.7}(year^{-1})$); and $s_0 e^{-\lambda_t t} s(N_p)$ represents the number of cells that arise from the thymus where $s(N_p)$ is the rate of export of the thymus defined by: $s(N_p) = \frac{1}{1+\frac{\bar{s}N_p}{N_p}}$. In (1), $\lambda_n N$ represents the naive cells' incorporation into the naive proliferating pool and $\lambda_n$ is the naive proliferation rate (which has the values 0.22, 2.1, 0.003, 0.005 in [8]). $\mu_n$ is the thymic naive cell death rate (equal to 4.4); $\mu_n g(N_p)N$ represents the naive cell death rate and the function $g(N_p)$ is the death rate of between naive TREC-positive and naive TREC-negative, defined as: $g(N_p) = 1 + \frac{\frac{bN_p}{N_p}}{1+\frac{N_p}{N_p}}$. $\bar{N}_p$ and $\bar{s}$ are equilibrium and scaling values respectively. The second differential equation is:

$$\frac{dN_p}{dt} = \lambda_n N + [ch(N, N_p) - \mu_n]N_p + \lambda_{mn}M, \quad (2)$$

where $c$ is the proliferation rate (with value 0 (no proliferation) or equal to $\mu_n(1 + \frac{300}{N_p})$); $ch(N, N_p)N_p$ represents the naive proliferation where $h(N, N_p)$ is the dilution of thymic-naive through proliferation defined by: $h(N, N_p) = \frac{1}{1+\frac{N+N_p}{N_p}}$. $\mu_n N_p$ is the death rate of proliferation-originated naive cells and $\lambda_{mn}$ is the reversion rate from memory into $N_p$, which has the values 0 and 0.5 in the experiments presented in [8]. The final differential equation is:

$$\frac{dM}{dt} = \lambda_a A - \mu_m M - \lambda_{mn}M, \quad (3)$$

where $\lambda_a$ is the rate of reversion of activated cells to memory cells and $\mu_m$ is the death rate of memory cells (equal to 0.05).

Initial SD simulations using equations (1) to (3) and approximated values for the initial size of each population of cells and the ratios that rule the kinetics of each population showed a decay in thymic output after the age of about twenty that became more pronounced with time (see Figure 2). This validate the results presented in [8]. Future work will use variations on the values for the ratio variables used in [8] in order to understand the importance of each individual integrant in the system. For example, it is important to establish how much the reversion of memory cells to a naive phenotype ratio impacts upon the depletion of naive T cells with age, the point in time that the system could be defined as losing functionality, and the impact of the naive proliferation rate on the final simulation results.

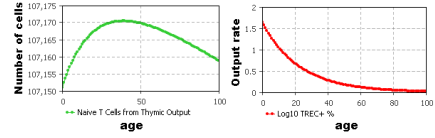

Figure 2: Decay of naive T cells from Thymus.

Studies on age-related declines in CD4+ T cell function [7] show that there is a reduced T-cell receptor signaling intensity, defects in activation, differentiatiation and expansion after stimulation, reduced production of interleukin-2 and impaired ability to provide cognate help to B cells after immunization. These factors can be explained by the fact that, without a continuous stream of thymic output, T cells have increased longevity to compensate for the depletion of new incoming cells. Several studies support the fact that this forced longevity of T cells is sufficient for the accumulation of age-related CD4+ T cells defects. Therefore, population-based models of T-cell repertoire evolution could guide new developments for treating diseases such as autoimmunity and helping the recovery of the system after depletion caused by infections, radiation and age. Future work will aim to develop the SD model into an ABS model to simulate components, functions and interactions involving the immunosenescence of T cells, taking the aspects cited above into consideration. This should permit the replication of immuno-responses to stimuli at the cellular level over the course of a lifetime and the simulation of thymic reconstitution with benefits and problems associated.

## 6   Conclusions

Understanding immunosenescence and its causes may help to reverse some of its consequences and improve life expectancy. Here, the important factors related to immunosenescence have been identified as the shrinkage of the naive T and B cell repertoire, decrease of innate immune cell diversity, filling of the immunological space by only a few types of phenotypical clones, accumulation of memory cells, loss of clones, inflammation, excessive T cell suppression, degeneration, auto-immunity and a decrease in response to vaccination. One of the main factors identified as influent in the process of immunosenescence is the number and phenotypical variety of naive T cells in an individual, which changes with age in quantity and diversity. At the beginning of life, the thymus is the principal source of naive T cells. With age, there is a decay in thymus output and a shift between the main source of naive T cells. It is believed that the sustaining of naive T cells in the organism is provided by peripheral expansion, reversion from a memory phenotype, and long-lived T cells.

This work has presented a review of the immunosenescence phenonmenon and has proposed extending an existing SD model of the processes involving the maintenance of the naive T cell repertoire into an ABS model. This should provide insight into how much each of the naive T cell sustaining possibilities influences the final pool of T cells. Initial experiments with the SD model have supported the the notion of thymic decay beginning at about age twenty and becoming more pronounced with age, but the use of an ABS model would permit the replication of immuno-responses to stimuli at the cellular level over the course of a lifetime, providing further insights into the immunosenescence phenomenon. The ultimate goal of the ABS model is to facilitate sufficient understanding of aging processes so that boosting techniques for other real-world degenerative systems may be developed.

## References

[1] A. Agrawal, S. Agrawal, J. Tay, and S. Gupta. Biology of dendritic cells in aging. *J Clin Immunol*, 28:14–20, 2007.

[2] M. Bulatti, M. Pellican, S. Vasto, and G. Colonna-Romano. Understanding ageing: Biomedical and bioengineering approaches, the immunologic view. *Immunity & Ageing*, 5(9), september 2008.

[3] S. Butcher, H. Chahel, and J. M. Lord. Ageing and the neutrophil: no appetite for killing? *Immunology*, 100:411–416, 2000.

[4] G. Candore, G. Colonna-Romano, C. R. Balistreri, D. D. Carlo, M. P. Grimaldi, F. List, D. Nuzzo, S. Vasto, D. Lio, and C. Caruso. Biology of longevity: Role of the innate immune system. *Rejuvenation Research*, 9(1):143–148, 2006. PMID: 16608411.

[5] C. Franceschi, M. Bonaf, and S. Valensin. Human immonosenescence: the prevailing of innate immunity, the failing of clonotypic immunity, and the filling of immunological space. *Vaccine*, 18:1717–1720, 2000.

[6] M. D. Martinis, C. Franceschi, D. Monti, and L. Ginaldi. Inflamm-ageing and lifelong antigenic load as major determinants of ageing rate and longevity. *FEBS*, 579:2035–2039, february 2005.

[7] A. C. Maue, E. J. Yager, S. L. Swain, D. L. Woodland, M. A. Blackman, and L. Haynes. T-cell immunosenescence: lessons learned from mouse models of aging. *Trends in Immunology*, 30(7):301–305, July 2009.

[8] J. M. Murray, G. R. Kaufmann, P. D. Hodgkin, S. R. Lewin, A. D. Kelleher, M. P. Davenport, and J. Zaunders. Naive t cells are maintained by thymic output in early ages but by proliferation without phenotypic change after twenty. *Immunology and Cell Biology*, (81):487–495, June 2003.

[9] S. Sharma, A. L. Dominguez, and J. Lustgarten. High accumulation of t regulatory cells prevents the activation of immune responses in aged animals. *The Journal of Immunology*, 177:8348–8355, september 2006.

[10] D. Wu and S. N. Meydani. Mechanism of age-associated up-regulation in macrophage pge2 synthesis. *Brain, Behavior, and Immunity*, 18(6):487 – 494, 2004.